\documentclass[runningheads]{llncs}
\usepackage[T1]{fontenc}
\usepackage[margin=1in]{geometry}
\usepackage{booktabs}
\usepackage{amsmath}
\usepackage{amssymb}
\usepackage{graphicx}
\usepackage{multirow}
\usepackage{subcaption}
\usepackage{xcolor}
\usepackage{hyperref}
\usepackage{tabularx}
\usepackage{enumitem}
\usepackage{algorithm}
\usepackage{algorithmic}
\usepackage{microtype}
\usepackage{parskip}
\usepackage{caption}

\hypersetup{colorlinks=true,linkcolor=blue!70!black,citecolor=blue!70!black,urlcolor=blue!70!black}

\title{\textbf{Concept-Constrained Prompt Learning for\\Few-Shot CLIP Adaptation}}
\author{ Na Sang\inst{1} \and Ding Ma\inst{2} \and Rui Sang\inst{3} \and Yuxuan Liu\inst{3} } \authorrunning{N. Sang et al.} \institute{ University of California, San Diego, La Jolla, CA, USA \and Georgia Institute of Technology, Atlanta, GA, USA \and Independent Researcher }
\authorrunning{N. Sang et al.}
\date{}

\begin{document}
\maketitle

\begin{abstract}
Few-shot prompt learning is an effective strategy for adapting CLIP to downstream tasks,
but class-only prompt optimization can overfit base-class supervision and weaken transfer
to unseen classes.
We propose \textbf{Concept-Constrained Prompt Learning (CCPL)}, a lightweight
regularization framework that anchors learnable class prompts to frozen
concept-level text prototypes without updating CLIP encoders.
CCPL learns a set of shared context tokens, instantiates class prompts by appending
class names, and constructs frozen concept prototypes from a class-level concept bank.
During training, a text-space cosine consistency objective aligns learnable class-prompt
embeddings with frozen concept prototypes; concept dropout provides additional
regularization against over-reliance on fixed concept lists.
At inference, CCPL optionally fuses class-prompt logits with concept-prototype logits
using a controllable ensemble weight~$\alpha$.
Our default configuration uses text-space concept regularization
($\lambda_{\mathrm{text}}=0.5$, concept dropout $p=0.3$) and weak concept-guided
fusion ($\alpha=0.1$), with no KL-based prediction consistency term.
Experiments under identical automatically-generated fallback splits show that
CCPL improves the base-to-new harmonic mean on DTD (+0.6) and EuroSAT (+2.9)
compared with CoOp, while remaining near-neutral on OxfordPets ($-$0.1).
Ablations indicate that text-space concept regularization is consistently beneficial,
while the best concept-guided inference strength is dataset- and protocol-sensitive.
These results suggest concept constraints are most effective when concept prototypes
align naturally with dataset semantics, and identify fine-grained categories as
a current boundary condition. The code is released at: https://github.com/richael-sang/concept-constrained-prompt-learning.

\end{abstract}

\section{Introduction}

CLIP-style vision-language pretraining provides a strong foundation for transferable
visual recognition by aligning image and text representations in a shared embedding
space~\cite{clip}.
The success of CLIP stems from training on massive image-text pairs, which instills
broad semantic knowledge about visual concepts across diverse domains.
When adapting CLIP to specific downstream tasks in low-data regimes, a core challenge
is how to specialize the model while preserving its open-vocabulary generalization capacity.

Prompt learning has become a practical adaptation strategy for CLIP:
instead of fine-tuning the entire model, methods such as CoOp optimize a small
number of learnable context tokens prepended to class names while keeping the
image and text encoders frozen~\cite{coop}.
This design is efficient and can substantially improve classification accuracy on
base classes with limited labeled examples.

However, few-shot prompt learning can still overfit base-class supervision.
In the standard base-to-new evaluation protocol~\cite{coop,cocoop}, models are trained
on a subset of \emph{base} classes with few-shot labels and evaluated on held-out
\emph{new} classes at test time.
This setting specifically probes whether adaptation preserves CLIP's
open-vocabulary semantic structure.
It is common to observe a trade-off: optimizing context tokens on base classes can
improve seen-class accuracy while degrading or failing to improve transfer to
unseen classes.
This tension arises because class-only prompt updates
(i.e., learning context tokens to maximize base-class performance) may drive
text embeddings away from the semantically rich directions originally encoded by
CLIP's text encoder.

One reason class-only prompts can overfit is that class names are semantically sparse.
A class name like ``braided'' or ``forest'' compactly identifies a category but provides
few explicit visual cues about what features to look for.
CLIP's zero-shot power partly comes from the rich alignment between visual features
and natural language descriptions that go beyond simple labels.
For many visual categories, especially textures and scenes, attributes such as
material, pattern, color, layout, and scene elements provide more descriptive
information than the class name alone.

If such descriptive concept phrases can be compiled into frozen text prototypes,
they can serve as semantic anchors in text space.
Aligning learnable class prompts toward these anchors would encourage the prompt
learning process to stay close to the semantic directions that CLIP associates with
meaningful visual attributes, rather than drifting toward dataset-specific biases.

We propose \textbf{Concept-Constrained Prompt Learning (CCPL)}, a framework that
introduces class-level concept phrases as frozen text-space anchors to regularize
prompt learning.
CCPL keeps CLIP's image and text encoders entirely frozen and adds no trainable
parameters beyond the shared prompt context tokens that already exist in CoOp-style
methods.
The concept prototypes are precomputed from a hand-crafted concept bank and are
never updated during training.
The regularization cost is therefore a lightweight text-space consistency objective
computed once per forward pass.

Concept dropout provides additional robustness: during training, a subset of
concept phrases is randomly masked per class, reducing over-reliance on any
specific concept and encouraging the model to align with the overall semantic
direction rather than individual concept phrases.

At inference, CCPL can optionally blend class-prompt logits and concept-prototype
logits with a tunable fusion weight, allowing the degree of concept guidance to be
adjusted without retraining.

Experiments on DTD, EuroSAT, and OxfordPets under identical fallback splits show a
dataset-dependent picture.
CCPL improves base-to-new harmonic mean on DTD ($+$0.6) and EuroSAT ($+$2.9)
compared with CoOp, with the EuroSAT gain primarily driven by improved new-class
accuracy (from 56.3\% to 60.6\%).
On OxfordPets, the result is near-neutral ($-$0.1~H), indicating that gains are
not universal across dataset types.
Ablations confirm that text-space concept regularization consistently contributes,
while the optimal inference-time fusion weight is protocol-sensitive.

\paragraph{Contributions.}
\begin{itemize}[leftmargin=*,itemsep=2pt]
    \item We formulate concept-constrained prompt learning as text-space semantic
          regularization for few-shot CLIP adaptation, requiring no additional
          trainable parameters beyond shared context tokens.
    \item We construct frozen concept prototypes from a class-level concept bank and
          align learnable class-prompt text embeddings with them through a cosine
          consistency objective (\S\ref{sec:method}).
    \item We introduce concept dropout for training-time robustness and controllable
          concept-guided inference at test time, with an explicit discussion of the
          base-new trade-off induced by the fusion weight.
    \item We provide controlled experiments on DTD, EuroSAT, and OxfordPets with
          main results, ablations, sensitivity analysis, and explicit identification
          of boundary conditions where concept constraints are and are not helpful
          (\S\ref{sec:experiments}).
\end{itemize}

\section{Related Work}

\subsection{Vision-Language Pretraining and CLIP Adaptation}

Large-scale vision-language pretraining has enabled strong zero-shot transfer across
a wide range of visual recognition tasks.
CLIP~\cite{clip} is a representative example: by training an image encoder and a
text encoder jointly on hundreds of millions of image-text pairs via contrastive
learning, it produces a shared embedding space that supports zero-shot classification
through text prompts.
At test time, class predictions are made by matching image features against
text features produced from class-name prompts such as ``a photo of a [CLASS]''.
This zero-shot capability is powerful but requires no class-specific labeled data,
which motivates efficient adaptation methods when task-specific labels are available.

Several parameter-efficient adaptation approaches have been proposed.
CLIP-Adapter~\cite{clipadapter} adds lightweight adapter layers on top of frozen
CLIP features and tunes only the adapters on few-shot labeled examples.
Tip-Adapter~\cite{tipadapter} constructs a key-value cache from the training set
and performs cache-based retrieval at inference, enabling training-free or
training-light adaptation.
These adapter-based approaches demonstrate that strong downstream performance can
be achieved without modifying the pretrained encoders.
Our method similarly keeps CLIP encoders frozen and introduces no learned components
beyond text-space context tokens.

\subsection{Prompt Learning for Vision-Language Models}

Prompt learning adapts CLIP by optimizing textual context tokens while keeping
encoders frozen.
CoOp~\cite{coop} introduces learnable context tokens prepended to class names and
optimizes them on few-shot base classes.
Although CoOp significantly improves base-class accuracy, its base-to-new
generalization can degrade because the learned context may overfit base-class patterns.
Co-CoOp~\cite{cocoop} addresses this by conditioning context tokens on each input image,
producing instance-conditional prompts that better generalize across classes.
However, instance-conditioning adds computation at inference.

MaPLe~\cite{maple} explores multi-modal prompt learning by jointly tuning context
tokens in both the vision and language branches of CLIP, enabling deeper
vision-language interaction.
PromptSRC~\cite{promptsrc} regularizes prompt learning by encouraging the learned
prompts to stay close to the pretrained text and image representations, using
self-regularization to prevent forgetting.
KgCoOp~\cite{kgcoop} incorporates general text knowledge to reduce overfitting by
minimizing the divergence between learned and hand-crafted text prompts.
ProGrad~\cite{prograd} uses gradient alignment to selectively update prompts in
directions consistent with the general CLIP objective, avoiding updates that conflict
with pretrained knowledge.

Visual prompt tuning (VPT)~\cite{vpt} explores learnable tokens in the vision
branch, demonstrating that prompt learning can be effective in either modality.
Our work is closest in spirit to regularization-based prompt learning:
rather than constraining the prompt update direction (as in PromptSRC and ProGrad),
we introduce an explicit external semantic anchor in text space derived from
human-interpretable concept phrases.

\subsection{Regularization for Prompt Generalization}

A recurring challenge in prompt learning is balancing adaptation to base classes
with preservation of CLIP's open-vocabulary capabilities.
Several methods approach this through regularization or auxiliary objectives.
ProGrad~\cite{prograd} projects gradients to avoid conflicting with pretrained
knowledge; PromptSRC~\cite{promptsrc} uses a frozen copy of the pretrained features
as a regularizer.
KgCoOp~\cite{kgcoop} leverages general hand-crafted text knowledge to anchor learned
prompts.

Our approach shares the motivation of anchoring learned prompts to stable semantic
references, but differs in its use of \emph{class-level concept phrases} as the anchor.
Rather than relying on a single hand-crafted template (e.g., ``a photo of a [CLASS]''),
we build a concept bank that contains multiple semantic cues per class, allowing the
anchor to represent richer intra-class semantic structure.
The alignment is enforced through a cosine consistency loss in text embedding space,
which is simple to implement and adds no learnable parameters.

\subsection{Concept-Based and Attribute-Based Recognition}

Incorporating human-interpretable concepts or attributes into visual recognition
has a long history.
Attribute-based recognition~\cite{lampert2009learning} uses binary attributes to
describe categories and enable zero-shot transfer.
Concept Bottleneck Models~\cite{koh2020concept} predict human-defined concepts as
an intermediate representation and use them to make final predictions, improving
interpretability and enabling concept-level interventions.
DCLIP~\cite{menon2023visual} and related work explore using descriptive language
generated by large language models (LLMs) to improve zero-shot classification,
showing that concept-rich text descriptions substantially improve visual recognition
beyond simple class names.

Our work is inspired by this line of research but operates in the few-shot prompt
learning setting rather than zero-shot.
We use a hand-crafted concept bank rather than on-the-fly LLM generation,
which avoids API dependencies and makes results more reproducible in controlled
experimental settings.
The concept phrases serve as frozen text-space anchors that guide prompt learning
rather than being used directly as classification templates.
This allows the learnable class prompts to benefit from concept-level semantic
regularization while retaining the flexibility of end-to-end prompt optimization.

\section{Method}
\label{sec:method}

\subsection{Overview}

\begin{figure}[t]
  \centering
  \includegraphics[width=0.92\linewidth]{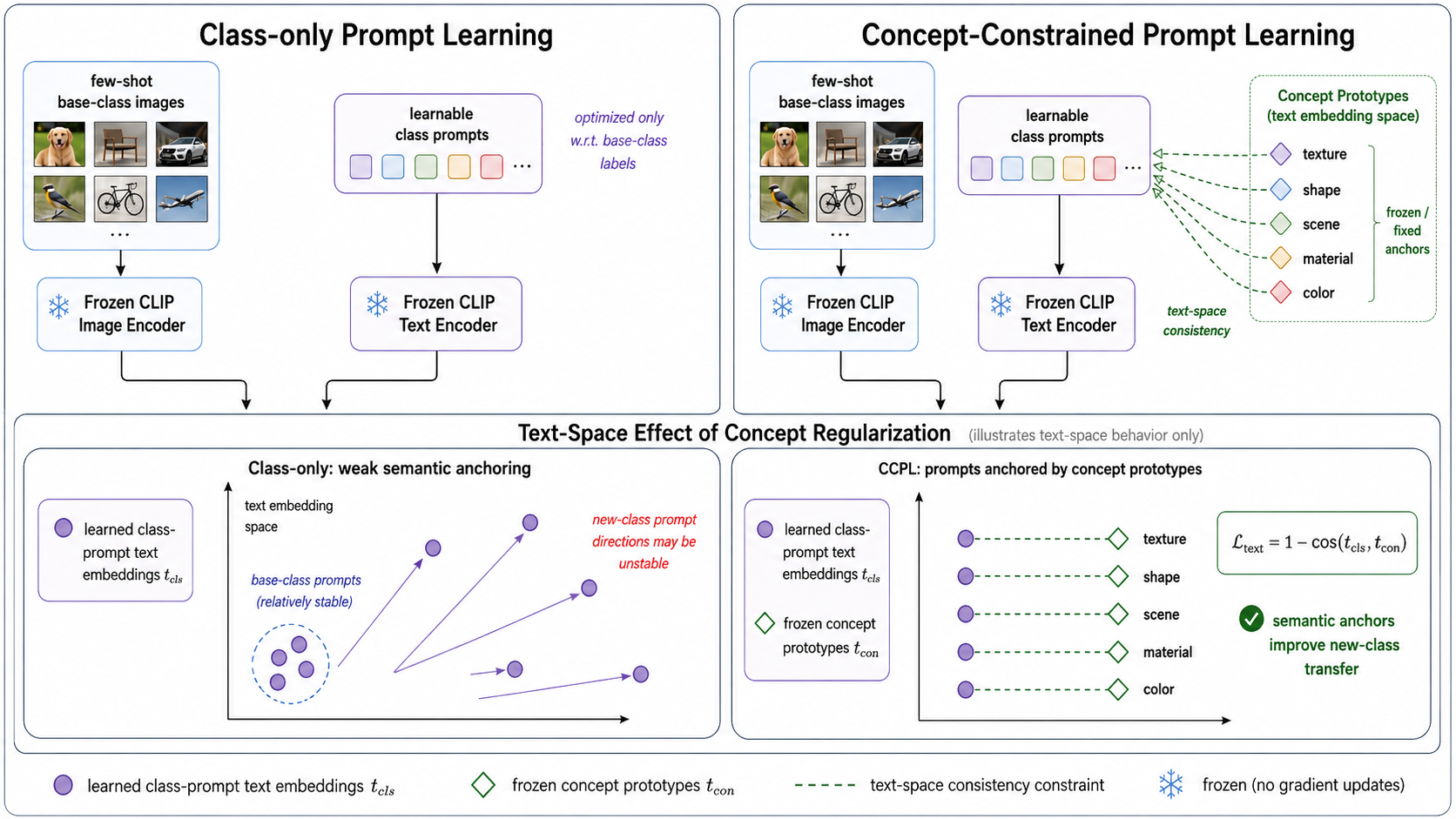}
  \caption{
    Overview of concept-constrained prompt learning (CCPL).
    Both class-only prompt learning and CCPL keep CLIP image and text encoders
    frozen and optimize prompt parameters using few-shot base-class supervision.
    CCPL additionally constructs frozen concept prototypes in text space and
    regularizes learned class-prompt text embeddings toward these prototypes
    using a cosine consistency objective.
    The schematic illustrates text-space behavior only and does not imply
    updates to frozen image embeddings.
  }
  \label{fig:ccpl_overview}
\end{figure}

CCPL introduces class-level concept phrases as frozen semantic anchors to regularize
prompt learning in CLIP.
Figure~\ref{fig:ccpl_overview} illustrates the overall framework.
Both CoOp and CCPL keep CLIP image and text encoders frozen and optimize
shared context tokens using few-shot base-class supervision.
CCPL additionally constructs frozen concept prototypes in text space and regularizes
learned class-prompt text embeddings toward these prototypes using a cosine
consistency objective.
At inference, CCPL can optionally blend class-prompt logits with concept-prototype
logits using a controllable fusion weight~$\alpha$.

\subsection{Problem Setup}

We consider few-shot adaptation of a pretrained CLIP model under a base-to-new protocol.
Let $\mathcal{C} = \mathcal{C}_{\mathrm{base}} \cup \mathcal{C}_{\mathrm{new}}$ be
the full class set, where $\mathcal{C}_{\mathrm{base}}$ and $\mathcal{C}_{\mathrm{new}}$
are disjoint.
Training uses a few-shot labeled set
$\mathcal{D}_{\mathrm{base}} = \{(x_i, y_i)\}_{i=1}^{N}$
over base classes only.
New classes are entirely unseen during training.
We keep CLIP image encoder $f_{\mathrm{img}}$ and text encoder $f_{\mathrm{text}}$
frozen throughout.
The only trainable parameters are the prompt context tokens.
Evaluation reports base accuracy, new accuracy, and harmonic mean:
\begin{equation}
H = \frac{2 \cdot \mathrm{Base} \cdot \mathrm{New}}{\mathrm{Base} + \mathrm{New}}.
\end{equation}

\subsection{Class Prompt Parameterization}

Following the CoOp-style parameterization~\cite{coop},
we learn a set of $M$ shared context tokens
$\theta = \{v_1, v_2, \ldots, v_M\}$
in the embedding space of the CLIP text encoder.
The class prompt for class~$c$ is formed by appending the class name token(s) to
the learned context:
\begin{equation}
p^{\mathrm{cls}}_c(\theta) = [v_1, v_2, \ldots, v_M, \mathrm{name}(c)],
\end{equation}
where $\mathrm{name}(c)$ is the tokenized class name and
the context tokens $v_1,\ldots,v_M$ are shared across all classes.
This means the same learned context is applied to both base and new classes:
the context is optimized on base classes and then directly applied to new classes
at evaluation, enabling generalization without class-specific parameters.

For image $x_i$, the frozen image encoder gives:
\begin{equation}
\mathbf{v}_i = \frac{f_{\mathrm{img}}(x_i)}{\|f_{\mathrm{img}}(x_i)\|_2}.
\end{equation}
For class $c$, the learnable class prompt text embedding is:
\begin{equation}
\mathbf{t}^{\mathrm{cls}}_c(\theta) =
  \frac{f_{\mathrm{text}}\!\left(p^{\mathrm{cls}}_c(\theta)\right)}
       {\left\|f_{\mathrm{text}}\!\left(p^{\mathrm{cls}}_c(\theta)\right)\right\|_2}.
\end{equation}
Class-prompt logits are computed via the standard CLIP cosine-similarity formula:
\begin{equation}
\mathbf{z}^{\mathrm{cls}}_i = s \cdot \mathbf{v}_i \left(\mathbf{T}^{\mathrm{cls}}\right)^\top,
\end{equation}
where $s = \exp(\log\_\mathrm{scale})$ is the CLIP logit scale and
$\mathbf{T}^{\mathrm{cls}} = [\mathbf{t}^{\mathrm{cls}}_1, \ldots, \mathbf{t}^{\mathrm{cls}}_C]$.

\subsection{Concept Bank and Concept Prototype Construction}

For each class $c$, we define $K_c$ concept prompts
$\{p^{\mathrm{con}}_{c,k}\}_{k=1}^{K_c}$
sourced from a hand-crafted concept bank (described below).
Each concept prompt encodes a descriptive attribute phrase associated with class~$c$,
for example, ``a photo of a forest, which contains dense tree canopy.''

Using the frozen CLIP text encoder, we compute normalized embeddings for each
concept prompt and average them to form a class-level concept prototype:
\begin{align}
\tilde{\mathbf{t}}^{\mathrm{con}}_c &=
  \frac{1}{K_c} \sum_{k=1}^{K_c}
  \frac{f_{\mathrm{text}}(p^{\mathrm{con}}_{c,k})}
       {\|f_{\mathrm{text}}(p^{\mathrm{con}}_{c,k})\|_2}, \\
\mathbf{t}^{\mathrm{con}}_c &=
  \frac{\tilde{\mathbf{t}}^{\mathrm{con}}_c}
       {\|\tilde{\mathbf{t}}^{\mathrm{con}}_c\|_2}.
\end{align}

\paragraph{Concept bank.}
The concept bank is a JSON file that maps dataset names and class names to
lists of concept phrases.
For EuroSAT (satellite scene classification), we provide manually authored
class-specific concepts that describe scene-level attributes:
for example, the class ``forest'' maps to phrases such as
``dense tree canopy,'' ``woodland area,'' and ``green vegetation.''
For DTD (texture classification), we use a template-based fallback:
each class uses five template phrases of the form
``\{class\_name\} texture,''
``\{class\_name\} surface pattern,'' etc.
For OxfordPets (fine-grained pet breed classification), template-based concepts
are similarly used.

An example of concept assignments across datasets is shown in
Table~\ref{tab:concept_examples}.
Class-specific concepts are more informative for scene and texture datasets,
where semantic attributes are salient and relatively generic.
For fine-grained categories, template-based concepts may only partially capture
breed-specific visual cues.

\paragraph{Scope of concept prototypes.}
The concept bank covers all classes in the dataset, including both base and new classes.
However, we emphasize that concept prototypes are computed \emph{solely from class names
and textual concept phrases}, without using any training images or labels from new
classes.
At training time, the text-space consistency loss $\mathcal{L}_{\mathrm{text}}$
is computed over base classes only, since only base-class prompts are optimized.
At inference, concept prototypes for all classes (including new classes) are used
to compute optional concept logits, which is semantically equivalent to CLIP's
standard zero-shot classification using descriptive text templates.

\subsection{Concept Dropout}

To reduce over-reliance on specific concept phrases and improve training-time
robustness, we apply concept dropout.
During training, each concept phrase for class~$c$ is retained with probability
$(1 - p_{\mathrm{drop}})$ independently:
\begin{equation}
m_{c,k} \sim \mathrm{Bernoulli}(1 - p_{\mathrm{drop}}),
\quad m_{c,k} \in \{0, 1\}.
\end{equation}
To ensure at least one concept is always retained, if all $m_{c,k} = 0$,
one concept is randomly selected and set to~1.
The dropout-masked concept prototype is then:
\begin{align}
\tilde{\mathbf{t}}^{\mathrm{con},\mathrm{drop}}_c &=
  \frac{1}{\sum_k m_{c,k}}
  \sum_{k=1}^{K_c} m_{c,k}
  \frac{f_{\mathrm{text}}(p^{\mathrm{con}}_{c,k})}
       {\|f_{\mathrm{text}}(p^{\mathrm{con}}_{c,k})\|_2}, \\
\mathbf{t}^{\mathrm{con},\mathrm{drop}}_c &=
  \frac{\tilde{\mathbf{t}}^{\mathrm{con},\mathrm{drop}}_c}
       {\|\tilde{\mathbf{t}}^{\mathrm{con},\mathrm{drop}}_c\|_2}.
\end{align}
At inference, we use the full concept set without dropout to form stable prototypes.
We use $p_{\mathrm{drop}} = 0.3$ by default.

\subsection{Text-Space Concept Regularization}

CCPL encourages the learnable class-prompt text embeddings to align with the frozen
concept prototypes through a cosine consistency loss.
Let $\mathbf{t}^{\mathrm{con}}_c$ denote the (possibly dropout-masked) concept
prototype.
The text-space consistency loss is:
\begin{equation}
\mathcal{L}_{\mathrm{text}} =
  \frac{1}{|\mathcal{C}_{\mathrm{base}}|}
  \sum_{c \in \mathcal{C}_{\mathrm{base}}}
  \left(1 - \cos\!\left(\mathbf{t}^{\mathrm{cls}}_c(\theta),\,
                          \mathbf{t}^{\mathrm{con}}_c\right)\right),
\end{equation}
where the sum is over base classes only, since those are the classes whose prompts
are being optimized during training.
This loss is zero when class-prompt embeddings are perfectly aligned with concept
prototypes and increases as they diverge in direction.
The standard cross-entropy loss on base-class predictions is:
\begin{equation}
\mathcal{L}_{\mathrm{ce}} = \mathrm{CE}\!\left(\mathbf{z}^{\mathrm{cls}}, y\right).
\end{equation}

\subsection{CCPL Training Objective}

Our default CCPL objective is:
\begin{equation}
\mathcal{L}_{\mathrm{CCPL}} =
  \mathcal{L}_{\mathrm{ce}} + \lambda_{\mathrm{text}} \mathcal{L}_{\mathrm{text}},
\label{eq:ccpl_loss}
\end{equation}
where $\lambda_{\mathrm{text}}$ controls the strength of concept regularization.
We use $\lambda_{\mathrm{text}} = 0.5$ by default.

\paragraph{Preliminary experiments with prediction-level consistency.}
We also investigated a KL-divergence loss between class-prompt logit distributions
and concept-prototype logit distributions during preliminary experiments.
However, this additional term was found to be less stable than text-space
regularization and did not consistently improve performance in our setting.
Therefore, the default CCPL configuration excludes it.
The default objective (Eq.~\ref{eq:ccpl_loss}) thus consists solely of
cross-entropy classification loss and text-space concept regularization.

\subsection{Concept-Guided Inference}

At inference, CCPL supports optional concept-guided logit fusion.
The concept-prototype logits are:
\begin{equation}
\mathbf{z}^{\mathrm{con}}_i = s \cdot \mathbf{v}_i \left(\mathbf{T}^{\mathrm{con}}\right)^\top,
\end{equation}
where $\mathbf{T}^{\mathrm{con}} = [\mathbf{t}^{\mathrm{con}}_1,\ldots,\mathbf{t}^{\mathrm{con}}_C]$
uses full concept sets without dropout.
Final logits are a convex combination:
\begin{equation}
\mathbf{z}_i = (1-\alpha)\,\mathbf{z}^{\mathrm{cls}}_i + \alpha\,\mathbf{z}^{\mathrm{con}}_i,
\end{equation}
where $\alpha \in [0,1]$ controls the reliance on frozen concept prototypes.
A small~$\alpha$ keeps predictions close to the learned class-prompt branch;
a larger~$\alpha$ increases concept guidance.
We use $\alpha=0.1$ by default, which provides a small concept-guided correction
without overriding the learned prompt.
Our ablations show that this parameter induces a base-new trade-off: larger $\alpha$
tends to improve new-class accuracy at the cost of base-class accuracy
(see Figure~\ref{fig:alpha_sensitivity}).

\subsection{Training Algorithm and Complexity}

Algorithm~\ref{alg:ccpl} summarizes the CCPL training procedure.
CCPL introduces no new learnable parameters beyond the shared context tokens~$\theta$
that already exist in CoOp-style methods.
The trainable parameter count is $M \times d$ where $M=16$ is the context length
and $d=512$ is the CLIP text embedding dimension for ViT-B/16, giving
$16 \times 512 = 8192$ trainable parameters out of approximately 150 million total
CLIP parameters ($\sim$0.005\% of the model).

Concept prototypes are frozen text embeddings precomputed once before training and
never updated.
The additional training cost consists of:
(1) computing $\mathbf{t}^{\mathrm{cls}}_c$ via the text encoder at each step,
which is the same as in CoOp; and
(2) evaluating $\mathcal{L}_{\mathrm{text}}$, which requires cosine similarity
between two normalized vectors per class -- a negligible overhead.
At inference, computing $\mathbf{z}^{\mathrm{con}}$ requires one additional
matrix multiplication using precomputed concept prototype embeddings.

\begin{algorithm}[t]
\caption{Training CCPL with text-space concept regularization}
\label{alg:ccpl}
\begin{algorithmic}[1]
\REQUIRE Few-shot base set $\mathcal{D}_{\mathrm{base}}$, class names,
         concept bank, frozen CLIP encoders $f_{\mathrm{img}}, f_{\mathrm{text}}$
\REQUIRE $\lambda_{\mathrm{text}}$, $p_{\mathrm{drop}}$
\STATE Initialize shared context tokens~$\theta$
\STATE \textbf{Precompute} frozen concept prototypes
       $\{\mathbf{t}^{\mathrm{con}}_c\}$ for all classes $c$
\FOR{each training epoch}
  \FOR{each mini-batch $(x_i, y_i)$ from $\mathcal{D}_{\mathrm{base}}$}
    \STATE $\mathbf{v}_i \leftarrow f_{\mathrm{img}}(x_i) /
                              \|f_{\mathrm{img}}(x_i)\|_2$
    \STATE Instantiate class prompts $\{p^{\mathrm{cls}}_c(\theta)\}$
    \STATE $\mathbf{t}^{\mathrm{cls}}_c \leftarrow
             f_{\mathrm{text}}(p^{\mathrm{cls}}_c(\theta)) /
             \|f_{\mathrm{text}}(p^{\mathrm{cls}}_c(\theta))\|_2$
           \quad for base classes $c$
    \STATE Apply concept dropout to get $\hat{\mathbf{t}}^{\mathrm{con}}_c$
    \STATE $\mathbf{z}^{\mathrm{cls}}_i \leftarrow
             s \cdot \mathbf{v}_i\, (\mathbf{T}^{\mathrm{cls}})^\top$
    \STATE $\mathcal{L}_{\mathrm{ce}} \leftarrow
             \mathrm{CE}(\mathbf{z}^{\mathrm{cls}}, y)$
    \STATE $\mathcal{L}_{\mathrm{text}} \leftarrow
             \frac{1}{|\mathcal{C}_{\mathrm{base}}|}
             \sum_c \left(1 - \cos(\mathbf{t}^{\mathrm{cls}}_c,
             \hat{\mathbf{t}}^{\mathrm{con}}_c)\right)$
    \STATE $\mathcal{L} \leftarrow \mathcal{L}_{\mathrm{ce}} +
             \lambda_{\mathrm{text}}\,\mathcal{L}_{\mathrm{text}}$
    \STATE Update $\theta$ only (CLIP encoders remain frozen)
  \ENDFOR
\ENDFOR
\STATE \textbf{Inference}: compute $\mathbf{z}_i =
       (1-\alpha)\mathbf{z}^{\mathrm{cls}}_i + \alpha\mathbf{z}^{\mathrm{con}}_i$
       using full concept prototypes (no dropout)
\end{algorithmic}
\end{algorithm}

\section{Experiments}
\label{sec:experiments}

\subsection{Experimental Setup}

\paragraph{Protocol caveat.}
Our experiments use automatically generated fallback splits from the repository
protocol rather than the official Zhou et al.\ splits~\cite{coop}.
Therefore, our reported numbers should be interpreted as controlled within-protocol
comparisons rather than direct comparisons to published official-split results.
All methods evaluated in this work share the same class splits, training schedule,
CLIP backbone, and evaluation code.

\subsection{Datasets and Protocol}

We evaluate on three datasets representing different visual recognition challenges
(Table~\ref{tab:datasets}).
\textbf{DTD}~\cite{dtd} is a texture recognition dataset with 47 classes capturing
visual material properties such as braided, dotted, and porous patterns.
\textbf{EuroSAT}~\cite{eurosat} is a satellite image classification dataset with 10
scene classes including annual crops, forests, highways, and residential areas.
\textbf{OxfordPets}~\cite{oxford_pets} is a fine-grained animal breed dataset with
37 cat and dog breeds.

For all datasets, we use a base-to-new protocol with 4-shot training on base classes
and evaluate on both base and new classes.
We report base accuracy, new-class accuracy, and harmonic mean~$H$.
All models use CLIP ViT-B/16 as the frozen backbone.

\begin{table}[t]
\centering
\caption{Datasets and evaluation protocol used in this work.}
\label{tab:datasets}
\small
\begin{tabular}{@{}lllccc@{}}
\toprule
Dataset & Type & Protocol & Shots & Epochs & Metric \\
\midrule
DTD~\cite{dtd} & Texture & Base-to-new & 4 & 50 & Base/New/H \\
EuroSAT~\cite{eurosat} & Scene & Base-to-new & 4 & 50 & Base/New/H \\
OxfordPets~\cite{oxford_pets} & Fine-grained & Base-to-new & 4 & 50 & Base/New/H \\
\bottomrule
\end{tabular}
\end{table}

\subsection{Implementation Details}

All experiments use CLIP ViT-B/16 with image and text encoders frozen.
We use 16 context tokens initialized randomly (CSC=False, shared across all classes)
with class name appended at the end of the context.
Training uses SGD with learning rate 0.002, cosine learning rate schedule,
warmup epoch~1, and batch size~32.
Evaluation batch size is~100.

For CCPL-default: $\lambda_{\mathrm{text}}=0.5$, concept dropout $p=0.3$,
inference fusion $\alpha=0.1$.
The concept bank is loaded from a JSON file shipped with the code;
EuroSAT uses class-specific concepts while DTD and OxfordPets use template-based
concepts of the form ``\{class\_name\} texture'' / ``\{class\_name\} surface pattern'', etc.
All experiments use seed~1 unless noted; DTD robustness results include seed~2.
All training and evaluation were performed on NVIDIA RTX 4090 GPUs.

\subsection{Result Provenance and Reproducibility}

All numbers reported in the tables and figures are taken from experiment logs in
the project workspace.
The plotting script at \texttt{paper/scripts/make\_figures.py} first searches the
structured output directories under \texttt{outputs/} and extracts the final
\texttt{* accuracy: ...\%} line from the corresponding \texttt{log.txt} files.
If a log is absent, the script falls back only to values already reported in the
manuscript tables, with those fallback values explicitly marked in code comments.
No figure uses interpolated or synthetic values.

For the main base-to-new results, the relevant logs are the CoOp and CCPL
base/new evaluation directories under \path{outputs/base2new50/}.
For EuroSAT ablations, logs are loaded from the base and new evaluation folders
under \path{outputs/base2new_ablation/}.
For DTD seed robustness, the script reads both seed~1 and seed~2 evaluation logs
from the same \path{base2new50} result group.
This provenance is important because the experiments use automatically generated
fallback splits: reproducibility depends on keeping the same class split files,
seed values, training schedule, and evaluation code.

The reported harmonic means are recomputed from extracted base and new accuracies:
\[
H = \frac{2BN}{B+N},
\]
where $B$ and $N$ denote base and new accuracy.
This avoids relying on a separately logged harmonic-mean number and ensures the
same formula is used across all methods and datasets.
All generated figures are vector PDFs saved under
\texttt{paper/figures/generated/}, with PNG copies produced only for quick visual
inspection.

\subsection{Main Base-to-New Results}

\begin{table}[t]
\centering
\caption{Main base-to-new results under the identical fallback split protocol (4-shot, seed=1, 50 epochs).}
\label{tab:main_results}
\small
\begin{tabular}{@{}llccc@{}}
\toprule
Dataset & Method & Base & New & H \\
\midrule
DTD & CoOp & 72.3 & 55.2 & 62.6 \\
DTD & CCPL-default & 73.8 & 55.3 & \textbf{63.2} \\
\midrule
EuroSAT & CoOp & 79.8 & 56.3 & 66.0 \\
EuroSAT & CCPL-default & 79.7 & 60.6 & \textbf{68.9} \\
\midrule
OxfordPets & CoOp & 94.9 & 97.7 & \textbf{96.3} \\
OxfordPets & CCPL-default & 94.9 & 97.5 & 96.2 \\
\bottomrule
\end{tabular}
\end{table}

\begin{figure}[t]
  \centering
  \includegraphics[width=0.90\linewidth]{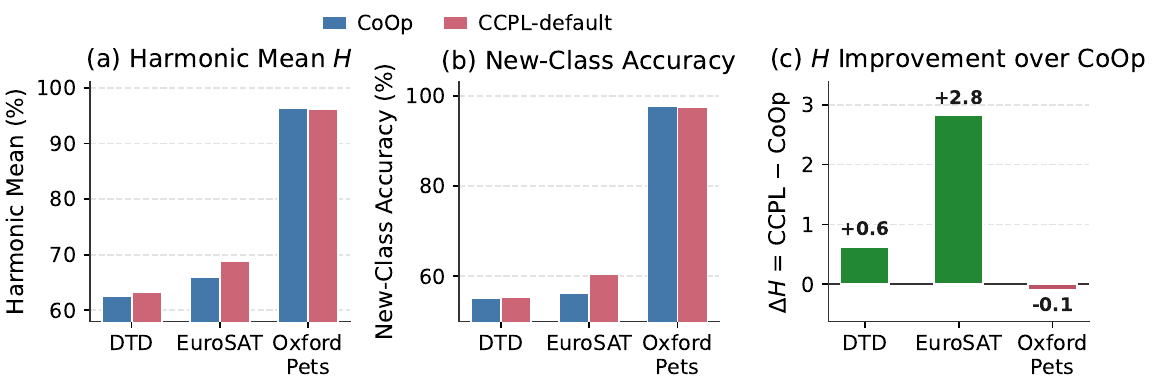}
  \caption{
    Main base-to-new results under the identical fallback split protocol.
    CCPL-default improves harmonic mean on DTD and EuroSAT, with the largest
    gain on EuroSAT driven by improved new-class accuracy.
    The near-neutral OxfordPets result highlights that concept constraints
    are not uniformly beneficial across fine-grained categories.
  }
  \label{fig:main_results}
\end{figure}

Table~\ref{tab:main_results} reports the main base-to-new results.
Figure~\ref{fig:main_results} visualizes the harmonic mean, new-class accuracy,
and per-dataset improvement over CoOp.

The largest improvement appears on EuroSAT, where the harmonic mean increases
from 66.0 to 68.9 ($+$2.9).
This gain is primarily driven by new-class accuracy, which increases from 56.3\%
to 60.6\% while base accuracy remains nearly unchanged (79.8 vs.\ 79.7).
This pattern is consistent with the hypothesis that concept regularization anchors
text embeddings closer to semantically rich directions that generalize to
unseen classes.

On DTD, CCPL-default achieves a moderate harmonic-mean improvement ($+$0.6),
with a small gain in base accuracy (72.3 to 73.8) and roughly equal new-class
accuracy (55.2 vs.\ 55.3).
The template-based DTD concepts (``\{class\} texture'') are simpler than
class-specific concepts, which may limit regularization strength.

On OxfordPets, CCPL-default achieves near-neutral performance relative to CoOp
($-$0.1 harmonic mean).
Base accuracy is unchanged (94.9) and new accuracy decreases slightly
(97.7 to 97.5).
OxfordPets is a high-accuracy setting where CoOp already achieves strong
generalization ($H=96.3$), leaving limited room for improvement.
The generic template-based concepts may not provide informative cues for
fine-grained breed discrimination.

\paragraph{Average harmonic mean.}
Averaged across the three datasets, CCPL-default achieves $H=76.1$
versus CoOp's $H=75.0$, an average gain of $+$1.1.
We emphasize that this average is dominated by the EuroSAT gain; the interpretation
should focus on per-dataset analysis rather than the aggregate number.

\subsection{Supporting Seed Robustness Evidence}

\begin{table}[t]
\centering
\caption{Additional available-seed and all-class results. These are supporting evidence rather than a full robustness study.}
\label{tab:supporting}
\small
\begin{tabular}{@{}llccc@{}}
\toprule
Setting & Dataset/Seed & CoOp & CCPL-default & Gain \\
\midrule
All-class & DTD (seed1) & 60.8 & 62.9 & +2.1 \\
All-class & EuroSAT (seed1) & 70.9 & 71.0 & +0.1 \\
All-class & DTD (seed2) & 60.1 & 60.9 & +0.8 \\
\midrule
Base-to-new H & DTD (seed1) & 62.6 & 63.2 & +0.6 \\
Base-to-new H & DTD (seed2) & 59.5 & 63.4 & +3.9 \\
\bottomrule
\end{tabular}
\end{table}

Table~\ref{tab:supporting} reports supporting results from a second seed (seed~2)
on DTD, which provides limited evidence of directional robustness.
These results should be interpreted as supporting evidence rather than a full
multi-seed evaluation.

\begin{figure}[t]
  \centering
  \includegraphics[width=0.72\linewidth]{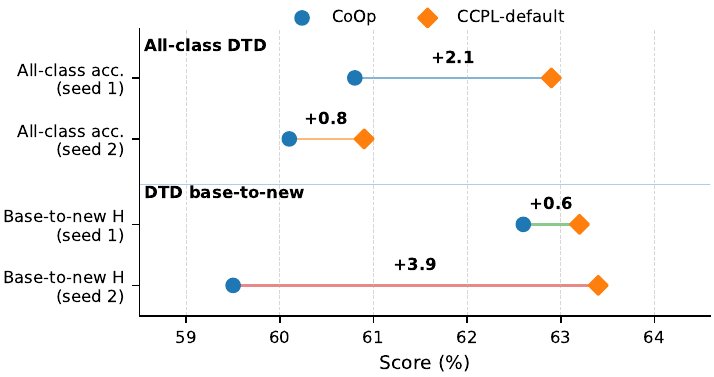}
  \caption{
    Robustness evidence from DTD seed~1 and seed~2.
    Although the current validation uses limited seeds,
    CCPL-default consistently improves over CoOp in the reported DTD settings.
  }
  \label{fig:seed_robustness}
\end{figure}

On DTD with seed~2, CCPL-default achieves $H=63.4$ compared with CoOp's $H=59.5$
($+$3.9).
The direction is consistent with seed~1, though the absolute numbers differ
(both CoOp and CCPL-default are lower under seed~2, and the gain is larger).
Figure~\ref{fig:seed_robustness} visualizes the per-seed comparison for DTD.

All-class few-shot results provide additional supporting evidence for DTD (seed~1:
$+$2.1, seed~2: $+$0.8) and EuroSAT (seed~1: $+$0.1), where CCPL-default
either gains or remains conservative relative to CoOp.

\subsection{Ablation Study}

\begin{table}[t]
\centering
\caption{EuroSAT base-to-new ablation under identical split protocol.}
\label{tab:ablation}
\small
\begin{tabular}{@{}lccccc@{}}
\toprule
Variant & $\lambda_{\text{text}}$ & $\alpha$ & Base & New & H \\
\midrule
CoOp & - & - & 79.8 & 56.3 & 66.0 \\
CCPL-default & 0.5 & 0.1 & 79.7 & 60.6 & 68.9 \\
no text regularization & 0.0 & 0.1 & 79.6 & 58.8 & 67.6 \\
no inference ensemble & 0.5 & 0.0 & 79.9 & 56.6 & 66.3 \\
stronger ensemble & 0.5 & 0.4 & 75.6 & 66.1 & \textbf{70.5} \\
\bottomrule
\end{tabular}
\end{table}

\begin{figure}[t]
  \centering
  \includegraphics[width=0.88\linewidth]{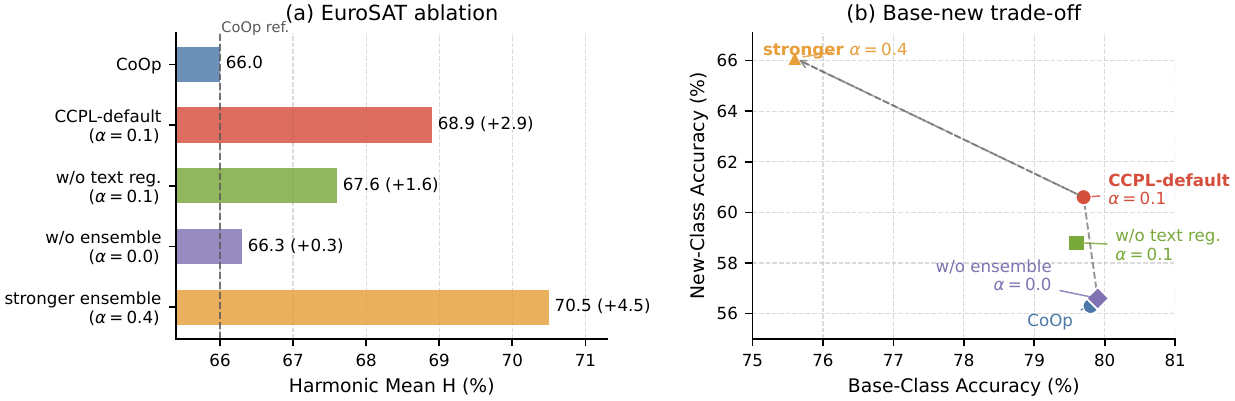}
  \caption{
    EuroSAT ablation and base-new trade-off.
    Panel~(a): harmonic mean for each ablation variant.
    Panel~(b): base-new accuracy scatter showing the trade-off controlled by~$\alpha$.
    Removing text-space regularization reduces $H$ relative to CCPL-default;
    removing concept-guided inference brings performance close to CoOp.
  }
  \label{fig:eurosat_ablation}
\end{figure}

We conduct an ablation study on EuroSAT to quantify the contribution of individual
CCPL components.
Figure~\ref{fig:eurosat_ablation} visualizes the ablation results and the base-new
trade-off.

\paragraph{Effect of text-space concept regularization.}
Removing text regularization ($\lambda_{\mathrm{text}}=0.0$, $\alpha=0.1$)
reduces harmonic mean from 68.9 to 67.6 ($-$1.3 vs.\ CCPL-default).
New-class accuracy drops from 60.6\% to 58.8\%, while base accuracy is nearly
unchanged (79.7 vs.\ 79.6).
This confirms that text-space concept regularization contributes to new-class
generalization, consistent with our hypothesis that aligning class-prompt
embeddings with concept prototypes preserves semantically transferable directions.

\paragraph{Effect of concept-guided inference.}
Removing inference ensemble ($\alpha=0.0$, $\lambda_{\mathrm{text}}=0.5$)
achieves $H=66.3$, close to CoOp's $H=66.0$.
New-class accuracy falls from 60.6\% to 56.6\%, suggesting that concept-guided
inference at $\alpha=0.1$ provides a meaningful additional boost beyond
what text regularization alone achieves.

\paragraph{Stronger concept ensemble.}
Using $\alpha=0.4$ achieves the highest harmonic mean of 70.5 on this protocol,
driven by a substantial new-class accuracy improvement to 66.1\%.
However, base-class accuracy decreases from 79.7 to 75.6.
This confirms that $\alpha$ controls a base-new trade-off: larger $\alpha$
increases reliance on frozen concept prototypes, which benefits new-class
transfer at the cost of base-class adaptation.

\subsection{Sensitivity to Concept-Guided Fusion Weight}

\begin{figure}[t]
  \centering
  \includegraphics[width=0.8\linewidth]{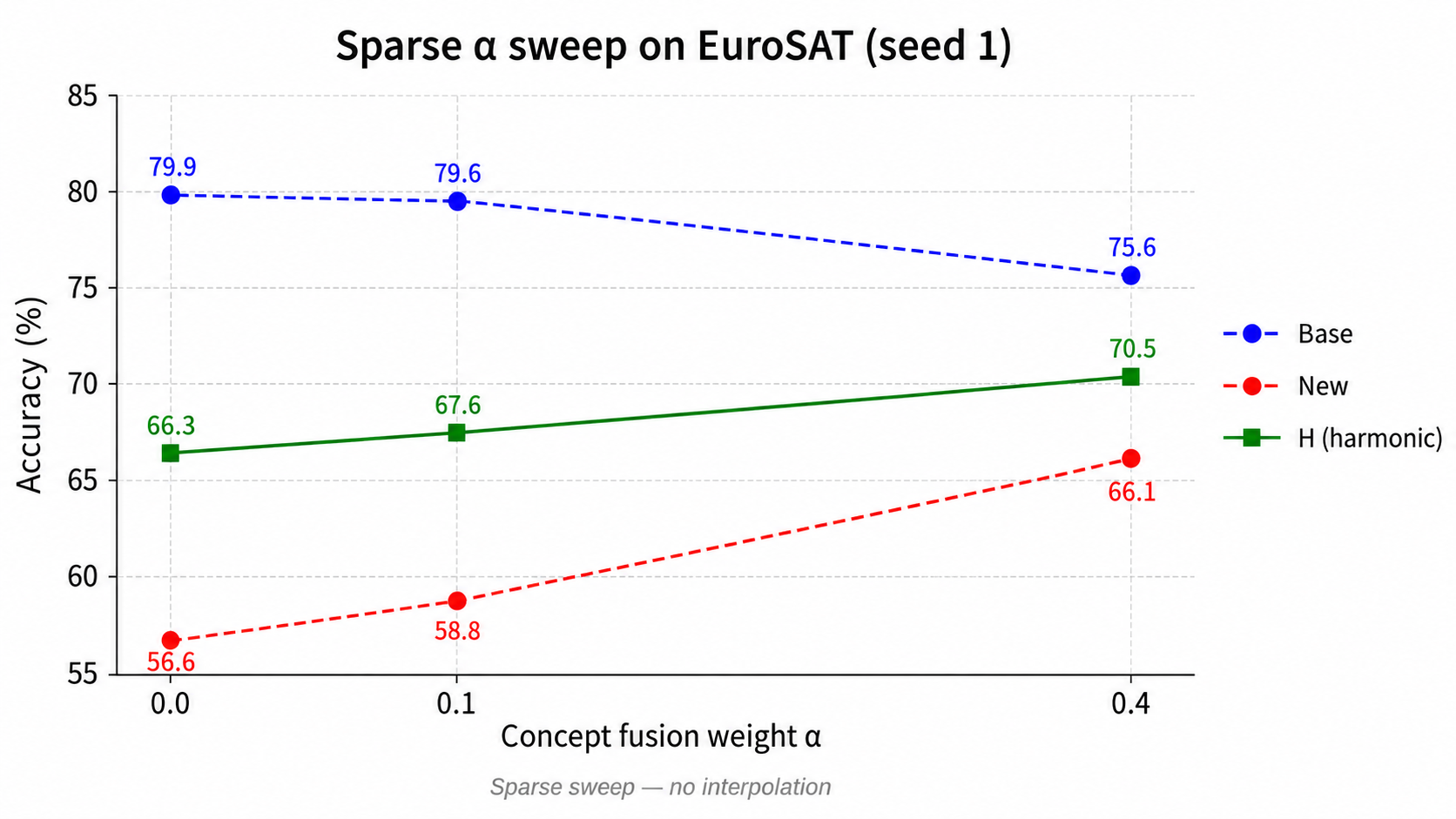}
  \caption{
    Sensitivity of concept-guided logit fusion weight~$\alpha$ on EuroSAT.
    A larger~$\alpha$ increases reliance on frozen concept prototypes.
    In this split, stronger fusion improves new-class accuracy but reduces
    base-class accuracy, suggesting that $\alpha$ should be treated as a
    controllable inference-time trade-off parameter rather than a universal constant.
    Only three values were evaluated; no interpolation is shown.
  }
  \label{fig:alpha_sensitivity}
\end{figure}

Figure~\ref{fig:alpha_sensitivity} shows the effect of the concept-guided fusion
weight~$\alpha$ on EuroSAT base-to-new performance.
Three values are evaluated: $\alpha \in \{0.0, 0.1, 0.4\}$.
As $\alpha$ increases, new-class accuracy monotonically increases while base-class
accuracy decreases, confirming the base-new trade-off.
The harmonic mean is maximized at $\alpha=0.4$ for this specific split, but
the base accuracy drops by approximately 4 points.

\paragraph{Important caveat.}
These results come from a single dataset and a single protocol.
The optimal $\alpha$ is likely sensitive to dataset semantics, split configuration,
and the quality of concept prototypes.
We therefore recommend treating $\alpha$ as an inference-time hyperparameter that
should be validated per dataset rather than a universal constant.
Our default of $\alpha=0.1$ is a conservative choice that keeps predictions
primarily determined by the learned class-prompt branch.

\subsection{Concept-Level Analysis}

Table~\ref{tab:concept_examples} lists representative concept prompts from
the CCPL concept bank for each dataset.
For EuroSAT, class-specific concepts describe scene-level attributes that naturally
correspond to visual patterns visible in satellite images.
For example, ``forest'' maps to ``dense tree canopy,'' ``woodland area,'' and
``green vegetation.''
These concepts capture salient scene-level cues that are likely to generalize
across different instances of the class.

For DTD, template-based concepts describe the texture class directly
(e.g., ``braided texture,'' ``braided surface pattern'').
These phrases are informative but less specific than class-tailored concepts.
The template approach means all DTD classes share the same concept structure,
with only the class name substituted.

For OxfordPets, the same template-based approach produces phrases like
``Siamese cat texture,'' which are not meaningful visual descriptions of the breed.
Fine-grained breed recognition typically requires discriminative cues about
specific shape, color patterns, and pose -- information that generic templates
cannot easily encode.
This suggests that the quality of concept prototypes is a key factor in whether
CCPL provides benefits, and motivates future work on dataset-specific concept
generation.

\begin{table*}[t]
  \centering
  \caption{
    Representative concept prompts from the CCPL concept bank.
    EuroSAT concepts capture scene-level semantics naturally, DTD concepts work well for regular texture names,
    while generic OxfordPets templates may miss fine-grained breed-specific cues.
  }
  \label{tab:concept_examples}
  \small
  \setlength{\tabcolsep}{5pt}
  \renewcommand{\arraystretch}{1.15}
  \begin{tabularx}{\textwidth}{@{}p{0.12\textwidth}p{0.14\textwidth}p{0.43\textwidth}X@{}}
    \toprule
    \textbf{Dataset} & \textbf{Class} & \textbf{Sample concept phrases} & \textbf{Utility / limitation} \\
    \midrule
    EuroSAT & forest
    & dense tree canopy; woodland area; green vegetation
    & Scene-level cues align naturally with class semantics. \\

    EuroSAT & highway
    & long paved road; linear transport; asphalt surface
    & Structural cues help distinguish road regions from surrounding areas. \\

    EuroSAT & sea lake
    & open water; lake surface; blue water region
    & Color and shape provide salient scene-level discriminators. \\

    DTD & braided
    & braided texture; braided surface pattern; braided visual texture
    & Template-based concepts capture the texture name directly. \\

    DTD & dotted
    & dotted texture; dotted surface pattern; dotted material pattern
    & Template-based concepts work well for regular visual patterns. \\

    OxfordPets & Siamese cat
    & Siamese cat texture; Siamese cat surface pattern; Siamese cat material pattern
    & Generic templates may miss subtle breed-specific cues. \\
    \bottomrule
  \end{tabularx}
\end{table*}

\section{Discussion and Limitations}

\subsection{When Do Concept Constraints Help?}

Our results reveal a consistent pattern: concept constraints are most effective
when the concept phrases capture salient visual attributes of the target dataset.

For EuroSAT, class-specific satellite scene concepts describe spatial and material
attributes that directly correspond to visually discriminative features
(vegetation density, water body characteristics, structural patterns).
These attributes are scene-level, relatively consistent across instances,
and well-captured by the CLIP text encoder's representation of natural language
descriptions.
The result is a clear improvement in new-class accuracy ($+$4.3 pp) with minimal
base-class impact.

For DTD, template-based texture concepts provide a weaker form of regularization
because they merely restate the class name with different surface forms
(``braided texture'' is semantically very close to the class name ``braided'').
Nevertheless, the small improvement in base accuracy ($+$1.5 pp) and stable
new-class accuracy suggest that even simple template-based anchoring provides
some structural benefit.

For OxfordPets, the generic template concepts are not informative for fine-grained
breed discrimination.
The discriminative information in this dataset (subtle differences in face shape,
fur color patterns, ear orientation) is typically encoded at a finer granularity
than template phrases like ``\{breed\} texture.''
In this setting, concept regularization may apply misaligned regularization:
pushing class-prompt embeddings toward uninformative concept directions can prevent
them from specializing to the fine-grained discriminative features.

\paragraph{Implications for concept design.}
These results suggest a practical guideline: concept constraints are most likely
to help when (1) the dataset involves scene-level or material-level semantics
(where generic visual attribute phrases are informative), and (2) the concept
phrases are specific enough to distinguish between classes.
Datasets with fine-grained visual categories and limited inter-class lexical
differences are more challenging for generic concept templates.

\subsection{The Role of Inference-Time Concept Guidance}

The ablation results show that concept-guided inference at $\alpha=0.1$ contributes
positively to EuroSAT performance ($H$ drops from 68.9 to 66.3 without ensemble).
This is notable because it suggests that frozen concept prototypes carry useful
classification information even without being used as a training objective.

The $\alpha$-sweep reveals a controllable base-new trade-off:
stronger concept guidance (larger $\alpha$) transfers more semantics from frozen
concept prototypes at the cost of overriding the learned base-class adaptation.
This is analogous to the trade-off in Tip-Adapter~\cite{tipadapter} between
cache-based retrieval and learned adapter features.

The sensitivity of the optimal $\alpha$ to the dataset and protocol means that
$\alpha$ should be treated as a per-dataset hyperparameter.
Our default $\alpha=0.1$ is a conservative choice appropriate for settings where
the concept alignment quality is unknown.

\subsection{Failure Modes and Boundary Conditions}

CCPL can fail or become neutral when concept prototypes are weakly aligned with
the visual decision boundary.
The OxfordPets result is an example: generic templates do not describe the subtle
breed-level distinctions needed for fine-grained recognition.
In such cases, the concept prototype may encode a broad semantic category
(``cat'' or ``dog'') rather than the fine-grained breed cues required for
classification.
Regularizing toward that prototype can then suppress useful task-specific
adaptation instead of improving transfer.

A second failure mode is concept ambiguity.
Some concept phrases may be visually meaningful but not class-discriminative.
For example, ``green vegetation'' is useful for EuroSAT forest and pasture
classes, but the phrase alone does not fully distinguish between them.
Uniformly averaging all concept phrases treats each phrase as equally informative,
which may dilute the prototype when some concepts are generic or shared across
classes.
This motivates learnable concept weighting or class-contrastive concept selection
in future work.

A third boundary condition concerns the class label space.
Our implementation can construct concept prototypes for all dataset classes,
including new classes, because class names are known at evaluation time.
This is consistent with standard CLIP zero-shot classification and base-to-new
prompt-learning protocols, where the label set is available.
However, it is not the same as an open-world discovery setting where new-class
names are unknown.
We therefore describe this as label-space semantic access rather than claiming
complete category ignorance.

\subsection{Limitations}

\paragraph{Limited evaluation scope.}
Our primary experiments cover three datasets (DTD, EuroSAT, OxfordPets) with
two seeds for DTD and one seed for EuroSAT and OxfordPets.
This is insufficient for a comprehensive evaluation of cross-dataset generalization.
Future work should extend to the full set of datasets used in CoOp and PromptSRC
evaluations (Caltech101, FGVCAircraft, Food101, Oxford Flowers, SUN397, UCF101,
ImageNet) with multiple seeds and official splits.

\paragraph{Comparison limited to CoOp.}
We compare primarily against CoOp rather than broader state-of-the-art
prompt-learning methods such as Co-CoOp, MaPLe, PromptSRC, KgCoOp, and ProGrad.
CCPL is implemented in the same framework as these methods, and future work should
provide fair comparisons using the same dataset splits and evaluation code.
We do not claim superiority over these methods; our results establish that
concept-constrained regularization can improve over the CoOp baseline in favorable settings.

\paragraph{Hand-crafted concept bank.}
The current concept bank is manually authored for EuroSAT class-specific concepts
and uses templates for DTD and OxfordPets.
This requires dataset-specific human effort and may not scale well.
An important direction is automatic concept generation using large language models
(LLMs) such as GPT or Llama, which could produce richer and more class-discriminative
concept descriptions without manual curation.

\paragraph{Fallback splits.}
Our experiments use automatically generated fallback splits rather than the official
Zhou et al.\ splits.
This means our absolute numbers cannot be directly compared with published results
from CoOp, Co-CoOp, MaPLe, and PromptSRC papers.
The comparisons in this paper are valid within our protocol but should not be
interpreted as improvements over those published baselines.

\paragraph{Concept alignment not directly measured.}
We do not directly measure how much CCPL changes the cosine alignment between
class-prompt embeddings and concept prototypes during training.
Logging text-space alignment dynamics would provide mechanistic evidence for
whether the regularizer achieves its intended effect, and is an important
diagnostic for future work.

\subsection{Future Work}

Several extensions are natural.
\textbf{Automatic concept generation:} using LLMs to generate class-specific
concept phrases for each dataset would eliminate manual curation and potentially
improve concept quality for fine-grained categories.
\textbf{Learnable concept weighting:} instead of uniform averaging over concept
phrases, learning per-class or per-concept weights could focus regularization on
the most informative concepts.
\textbf{Uncertainty routing:} dynamically routing predictions between the class-prompt
branch and the concept-prototype branch based on prediction confidence could
provide better adaptation in diverse settings.
\textbf{Official split evaluation:} re-running all experiments with official
Zhou splits and multiple seeds would strengthen the evidence base and enable
direct comparison with published baselines.
\textbf{Broader dataset coverage:} evaluating on more datasets, particularly
those with different semantic structures (object recognition, action recognition,
domain-specific categories), would characterize the scope of concept-constraint benefits.

\section{Conclusion}

We presented Concept-Constrained Prompt Learning (CCPL), a lightweight framework
for few-shot CLIP adaptation that regularizes learnable class prompts with frozen
concept-level text prototypes.
CCPL keeps CLIP encoders frozen, adds text-space concept consistency, uses concept
dropout for robustness, and supports optional concept-guided inference with
tunable strength.
Under identical fallback splits, CCPL-default improves base-to-new harmonic mean
on DTD and EuroSAT compared with CoOp, with EuroSAT gains mainly driven by
new-class accuracy.
On OxfordPets, performance is near-neutral with a slight decrease, indicating
that concept constraints are not universally beneficial across all dataset types.
Our ablations show that text regularization is useful in the reported EuroSAT
setting, while the best inference fusion strength depends on concept-class
alignment and evaluation protocol.
Overall, CCPL offers a practical and interpretable regularization direction for
prompt learning, especially for texture/scene-style transfer scenarios, and
highlights the importance of reporting boundary conditions rather than
overgeneralized claims.

\bibliographystyle{plain}
\bibliography{references}

\end{document}